\documentclass[letterpaper, 10 pt, conference]{ieeeconf}  

\IEEEoverridecommandlockouts                              

\usepackage{graphics} 
\usepackage{graphicx}
\usepackage{mathptmx} 
\usepackage{amsmath} 
\usepackage{amssymb}  

\graphicspath{ {./Figures/} }

\usepackage{algorithm}
\usepackage{algorithmic}
\usepackage{multirow}
\usepackage{tabularx}

\title{\LARGE \bf
Context-Aware Scene Prediction Network (CASPNet)
}

\author{Maximilian Schäfer$^{1}$, Kun Zhao$^{2}$, Markus Bühren$^{2}$ and Anton Kummert$^{1}$
\thanks{$^{1}$Maximilian Schäfer and Anton Kummert are with the School of Electrical, Information and Media Engineering,
        University of Wuppertal, 42119 Wuppertal, Germany {\tt\small  \{maximilian.schaefer, kummert\}@uni-wuppertal.de}}%
\thanks{$^{2}$Kun Zhao and Markus Bühren are with Aptiv Services Deutschland GmbH
        {\tt\small \{kun.zhao, markus.buehren\}@aptiv.com}}%
}

\begin{document}

\maketitle
\thispagestyle{empty}
\pagestyle{empty}

\begin{abstract}

	Predicting the future motion of surrounding road users is a crucial and challenging task for autonomous driving (AD) and various advanced driver-assistance systems (ADAS). Planning a safe future trajectory heavily depends on understanding the traffic scene and anticipating its dynamics. The challenges do not only lie in understanding the complex driving scenarios but also the numerous possible interactions among road users and environments, which are practically not feasible for explicit modeling. In this work, we tackle the above challenges by jointly learning and predicting the motion of all road users in a scene, using a novel convolutional neural network (CNN) and recurrent neural network (RNN) based architecture. Moreover, by exploiting grid-based input and output data structures, the computational cost is independent of the number of road users and multi-modal predictions become inherent properties of our proposed method. Evaluation on the nuScenes dataset shows that our approach reaches state-of-the-art results in the prediction benchmark.

\end{abstract}

\section{INTRODUCTION}

Anticipating road users' future motion is a key functionality to support various ADAS applications, such as collision avoidance, adaptive cruise control and lane change assistance. For fully autonomous driving, the prediction task plays an even more vital role. A feasible, safe and convenient path for the ego vehicle depends on a proper prediction of the traffic scenario. This is already a challenging task for highway applications, where the interaction among vehicles and the road structure affects the predictions significantly. It is an even more challenging task for urban applications. The variable number and types of road users with different maneuverabilities as well as complex traffic rules and variations of road structures make the prediction problem not feasible to model using conventional methods. Thus, data-driven approaches have gained focus in the last few years and made significant progress.

\begin{figure}[t!]
	\centering
	\includegraphics[width=0.39\textwidth]{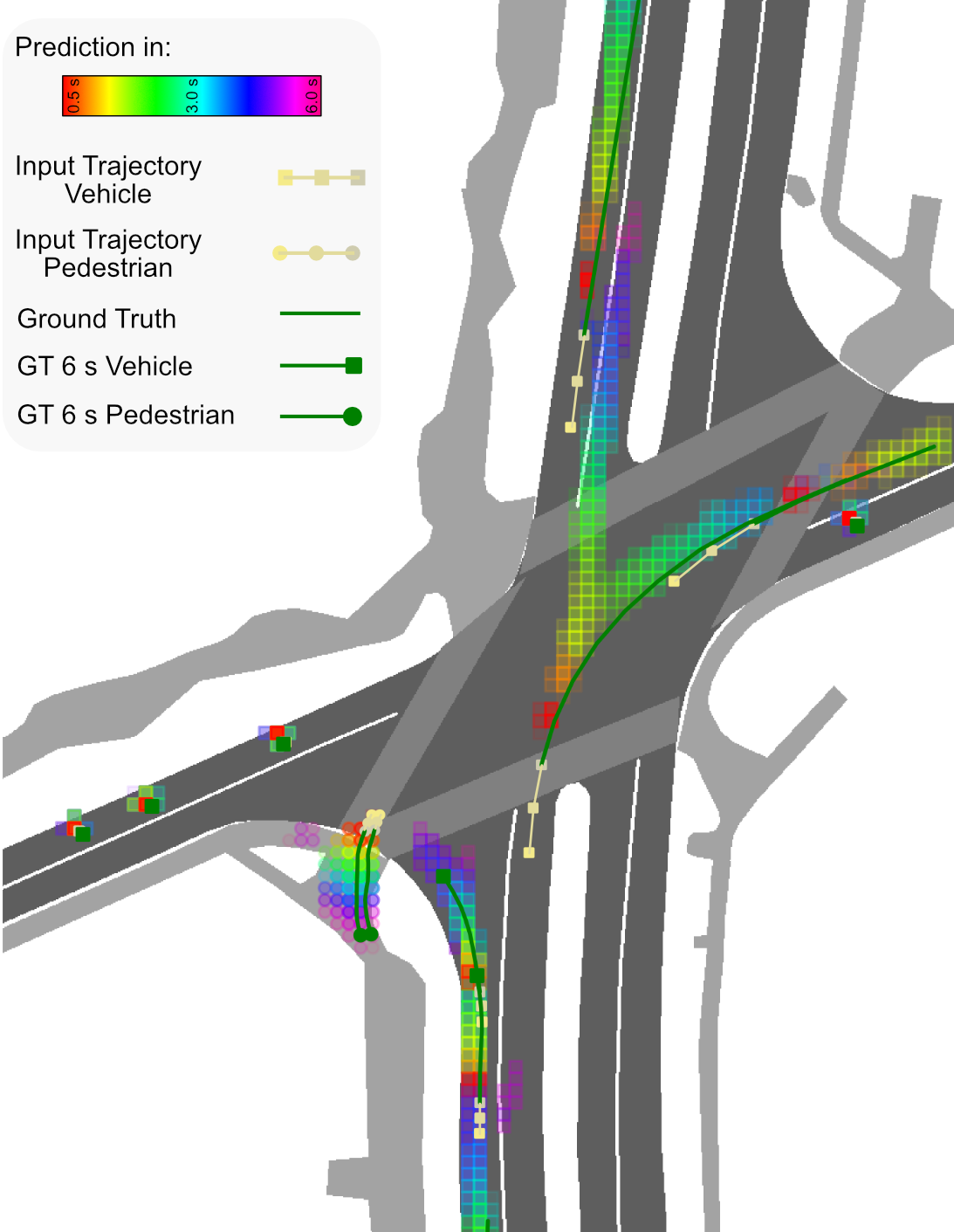}
	\caption{Example of CASPNet's prediction results (overlapped with the map for visual clarity) at an intersection in Singapore from the nuScenes dataset \cite{Caesar2020}. In this example, there are multiple vehicles and pedestrians, which move in different directions and face different environments. Circle and square points denote the trajectory points of vehicle and pedestrian respectively, in which yellow color represents the past trajectory and green color the future ground truth (GT). A heatmap color scheme is utilized to represent different prediction time horizons (red: 0.5 s to magenta: 6 s).}
	\label{fig_introduction}
\end{figure}

Following recent advancements in the field of motion forecasting, we propose the Context-Aware Scene Prediction Network (CASPNet), a network for jointly predicting all road users in a scene at the same time. To cope with the aforementioned challenges, we specifically designed a CNN- and RNN-based architecture for a multi-modal and context-aware prediction. To keep the complexity at a manageable size and to avoid mode collapse, we utilize a fixed-size grid-based input and output structure. Additionally, this allows the network to model distributions of road users' future positions without setting a predefined number of modalities. An attention module is introduced, to learn interactions among the road users as well as their static context at a variable range. We show the effectiveness of CASPNet for scene understanding and prediction on the challenging urban motion prediction data set nuScenes \cite{Caesar2020}. At the same time, for evaluation purpose we also applied it for single road user prediction, thus a fair comparison with state-of-the-art approaches can be achieved. Our contributions can be summarized as follows:

\begin{itemize}
	\item A novel network architecture is proposed to tackle the problem of predicting a variable number of road users, of different types, simultaneously in a context-aware and multi-modal manner. The computational cost of the proposed model is independent of the complexity of the traffic scene.
	\item We show an attention mechanism which learns interactions of road users at different ranges on 2D feature maps.
\end{itemize}

\section{RELATED WORKS}
Context-awareness and multi-modality are two crucial features for a state-of-the-art motion prediction system. They are key strengths of machine learning methods compared to traditional approaches, such as those listed in \cite{conventional}.

\subsection{Context-awareness}
For motion prediction, the term \textit{context} describes the traffic environment a road user is interacting with. We categorized context into two specific categories: \textit{dynamic context} and \textit{static context}. They have different characteristics and require different approaches. Dynamic context considers the interactions between road users. It is dynamic because each road user's behavior may affect others, and at the same time may be affected by others'. On the other hand, static context restricts the road users' behavior but not the other way around. Typical examples are road and lane structures, but also traffic lights, weather conditions and traffic rules.

One of the first approaches, which considers interactions among road users for motion prediction, is \cite{SocialLSTM}. The authors apply a long short-term memory (LSTM) network to iteratively encode each pedestrian's dynamic independently, then combine nearby pedestrians' hidden states through a "social pooling" operation. A similar concept is seen in \cite{Desire}. Following this concept, the authors of \cite{SocialGAN} used MLP and max pooling to fuse the movement information from multiple road users, such that the pooling operation is invariant to the number of neighboring road users. In \cite{Sophie}, the authors applied attention mechanisms \cite{visual-attention} to combine the LSTM hidden states. Different attention mechanisms, e.g. graph attention and multi-head attention, are seen in \cite{Graph-attention}, \cite{Messaoud2020}, \cite{girgis2021autobots} to integrate the dynamic context into the prediction systems.

The dynamic context comprises correlations among road users' motions. And, convolutional operations are good at learning the correlations among data under its kernel. This leads to the decision of using an image or image-like input data structure, so a CNN can be directly applied for learning the dynamic context in the scene. In \cite{Deo2018_soical} and \cite{Tensor-fusion}, an RNN is first used to encode each individual road users' features into latent codes. Then, these codes are arranged in a 2D grid map by filling the grid pixels, which match road users' positions with the codes. Due to increasing receptive fields in a CNN with increasing layer depth, the correlation among road users from near to far distances can be learned through multiple layers. When one considers the resolution of the grid map and maneuver capability of road users, an efficient design, regarding to depth of the CNN and each layer's parameters, can be achieved for specific application scenarios.

Similar approaches can be seen for integrating the static context. The static context is often represented as an image or image-like data structure and given to a CNN-based encoder for feature encoding. As examples, rasterized HD maps are used in \cite{Uber-paper}, \cite{Phan-Minh2020} and \cite{Bansal2019}. In \cite{Vectornet}, \cite{Khandelwal2020} the map information are simplified into either points, polygons or curves and encoded using graph neural networks (GNN).

\begin{figure*}[t]
	\centering
	\includegraphics[width=0.8\textwidth]{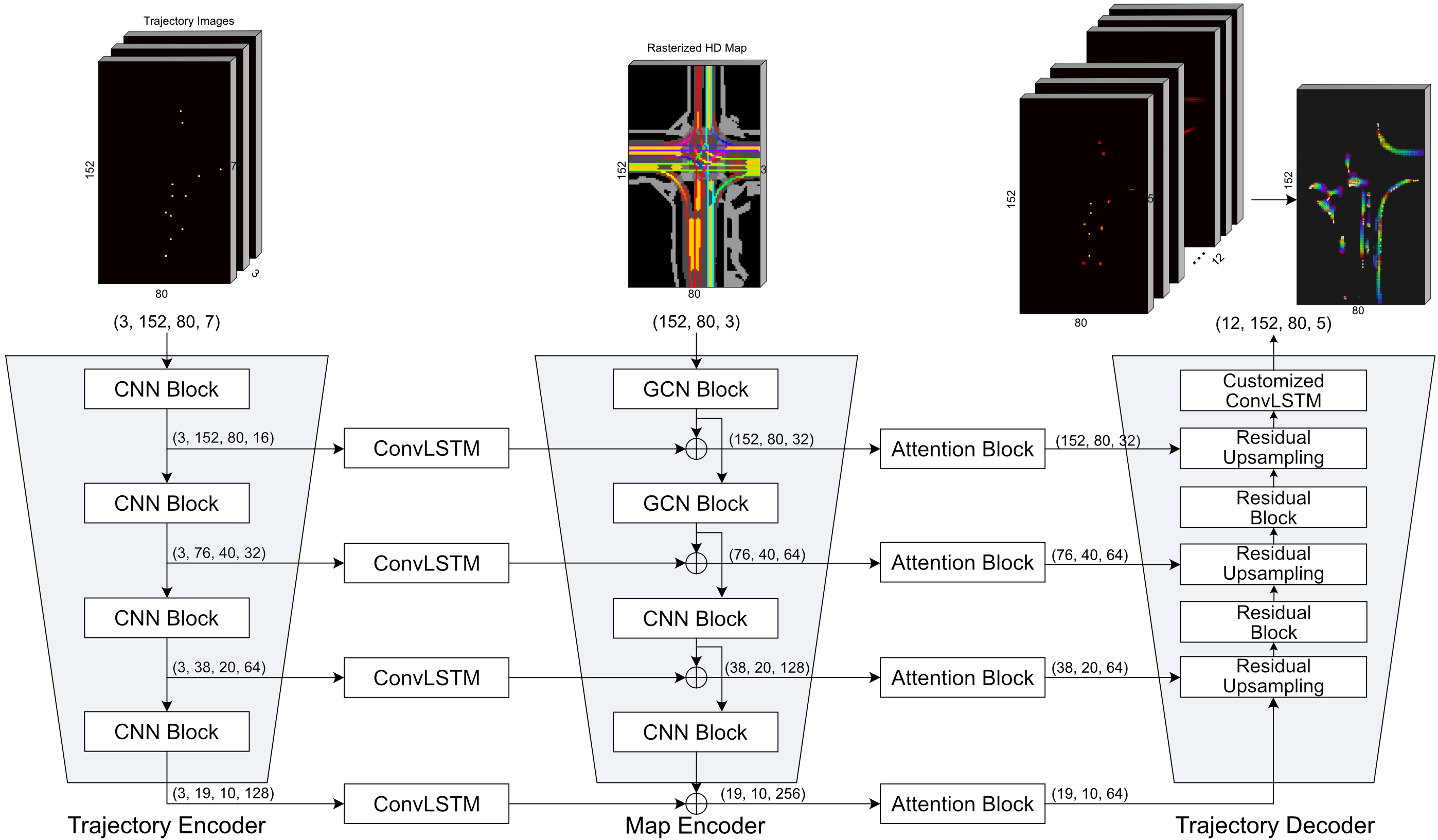} 
	\caption{Overview of CASPNet architecture as well as input and output definitions.}
	\label{fig_caspnet}
\end{figure*}

\subsection{Multi-modality}
Multi-modality means, there are a variable number of plausible future states given the same past motion and context. This is mainly because road users' driving intentions are not directly observable. Thus, a multi-modal prediction system needs to give different outputs for the same input trajectory. 

When using a parametric output representation to define the exact future position, multi-modal prediction requires the system either having a multi-head output design, or bringing randomness into the system to map the unknown modalities. In \cite{Multi-path}, \cite{Uber-paper}, \cite{Phan-Minh2020} and \cite{Ambiguity}, various architectures were introduced which all share a similar concept of multi-head outputs. This concept consequentially leads to a fixed number of modes. For training, the output closest to the ground truth trajectory is usually chosen for backpropagation \cite{Multi-path}, \cite{Ambiguity}. Furthermore, in \cite{Multi-path} the authors applied a Gaussian Mixture Model (GMM) to model the multi-modal output distribution. A slightly different approach was introduced in \cite{Deo2018_soical}, where the "teacher forcing" technique is used for conditional decoding based on the ground truth maneuver label during training. At testing time, all the possible maneuver labels can be repeatedly given to the decoder with the same code for conditioned decoding of different trajectory predictions.

Generative adversarial networks (GAN) \cite{GAN} or variational autoencoders (VAE) \cite{VAE}, are also commonly used for trajectory prediction tasks. Both GAN and VAE related approaches bring a randomness factor into an otherwise deterministic network, and utilize it to map the variable multi-modal output distributions. In the case of GAN, the noise input following a given distribution is sampled multiple times for the same training sample. This results in multiple network predictions, and the best among them is chosen for gradient backpropagation. This strategy is called "Minimum over N" in \cite{SocialGAN}. In systems that use VAE, a similar concept can be seen in \cite{Best-of-many}.

When using occupancy grid maps to model the possibility of future positions, the multi-modality property is inherently given, as shown in \cite{Prognosenet},  \cite{Rules-of-road} and \cite{Gilles2021}. Neither a special training strategy nor pre-labeling or clustering of the training data for multi-head outputs is necessary to facilitate diversity predictions. In this work, we fully utilized the advantages of a grid map representation to jointly predict all road users' future positions. The whole scene is predicted at once, instead of separate predictions for each individual road user.

\section{OUR APPROACH}

In this chapter, our system for jointly predicting the future trajectories of all road users in a scene is explained in detail. We start by defining the data structure and then present our network architecture as well as the loss function.

\subsection{Input and Output}
For the prediction of a scene, the future motion of all road users in a region of interest (ROI) is predicted. The number of road users may vary from scene to scene which leads to the requirement of storing a variable number of road users in a shared data structure.
As input and output data structure, we use a series of 2D grids (in the following also referred to as images) which allows the network to jointly encode the trajectories of all road users in an ROI and support the prediction of all road users. The static context input is defined as a three-channel image. Due to the fixed size input and output data structures, the computational cost of the proposed system is the same and independent of the number of road users in the scene and the complexity of the road or lane structures.

\subsubsection{Dynamic Context} 
The past trajectories of all road users in the scene consists of $M$ input time steps $T_{i}=(t_{-(M-1)},...,t_{-1}, t_{0})$. $t_0$ denotes the current frame at which the prediction is made. The past trajectories are rasterized into a series of input grids:
\begin{equation}
	I \in \mathbb{R}^{M \times U \times V \times |F_i|}
	\label{eq_input_structure}
\end{equation}
where $U$ and $V$ define the region surrounding the target vehicle and $|F_i|$ is the number of input features per pixel. We set the region size ($U$, $V$) to (152, 80) with a resolution of ($1 \times 1$) m per pixel. The number of past time step is set to ${M}=3$ with a temporal interval $\Delta_{t_i}=0.5$ s between consecutive frames. This leads to a maximum history of 1 s as our input data. The grids are aligned with the heading direction of the target vehicle at $t_0$. At time step $t_0$, the target vehicle is positioned at $u=122$ and $v=40$. Therefore, the ROI covers 122 m in front, 30 m behind and 40 m to the sides of the target vehicle. Other road users are rasterized relative to the target vehicle's position at $t_0$. In the $|F_i|$ image channels, we store the state of a road user at time step $t_i$:
\begin{equation}
	F_i=(c_{\text{target}}^{t_i}, c_{\text{vehicle}}^{t_i}, c_{\text{pedestrian}}^{t_i}, \delta_u^{t_i}, \delta_v^{t_i}, \textit{v}_u^{t_i}, \textit{v}_v^{t_i}).
	\label{eq_input_features}
\end{equation}
$c_{\text{target}}$, $c_{\text{vehicle}}$ and $c_{\text{pedestrian}}$ are one-hot encoded road user type information. Pedestrian and vehicle trajectories are currently utilized in our experiments. $c_{\text{target}}$ is optional, it is only utilized when comparing the prediction results with other works, more details in the evaluation section. To cover a bigger area while keeping the input feature maps small, we use a low resolution and describe the exact position within a pixel by defining an offset $\delta_u, \delta_v$ from the middle of a pixel. Moreover, we use the road user's velocity $\textit{v}$ in $u$ and $v$ direction as input.

\subsubsection{Static Context} We rasterize the content of a given HD map, similar to \cite{Djuric2020, Phan-Minh2020}, into a bird's eye view RGB image with a size of $(152, 80, 3)$, covering the same area as the input trajectory grid at $t_0$. Each semantic map layer type has a distinct color assigned to it. We successively raster starting with larger layers like driveable areas to smaller ones e.g. crossing areas. Additionally, the position and orientation of lane centerlines are encoded. Based on the orientation of a centerline at a specific point, a corresponding color from the HSV color space is assigned which is then converted back to RGB (e.g 0° resulting in red) like proposed in \cite{Djuric2020}.

\subsubsection{CASPNet Output}
As output $O$, we define a series of grids for the $N$ future time steps $T_{o}=(t_{1}, ..., t_N)$:
\begin{equation}
	O \in \mathbb{R}^{N \times U \times V \times |F_o|}
	\label{eq_output}
\end{equation}
where we set $N=12$ with a time interval $\Delta_{t_o}=0.5$ s. The region size stays consistent and the features per pixel are defined as follow:
\begin{equation}
	F_o=(c_{\text{target}}^{t_o}, c_{\text{vehicle}}^{t_o}, c_{\text{pedestrian}}^{t_o}, \delta_u^{t_o}, \delta_v^{t_o})
	\label{eq_output_features}
\end{equation}
where $t_o \in T_0$. Therefore, each pixel in the output grid describes the occupancy probabilities of one of the three classes at that location at future time step $t_o$. Similar to the input, $ \delta_u^{t_o} \text{ and } \delta_v^{t_o}$ describe the in-pixel offset.

\subsubsection{Ground Truth} The ground truth $Y \in \mathbb{R}^{N \times U \times V \times |F_o|}$ has the same structure as the previously described network output. Around a pixel where a road user's ground truth position is, a 2D Gaussian kernel is introduced to represent the occupancy uncertainty. The resulting Gaussian kernel is defined through standard deviations $\sigma_{long}$ (in the driving direction of that road user), $\sigma_{lat}$ (in the lateral direction) and an orientation $\Theta$. The standard deviations are dependent on the absolute velocity $\textit{v}^t$ of a road user at time step $t$ and an initial $\sigma_r^t$ which increases in time since we expect a higher uncertainty with an increasing time horizon and velocity:
\begin{equation}
	\sigma_{r}^t = {\frac{\sigma_{\text{max}, r}^t }{\textit{v}_{\text{max}}}} \textit{v}^t + \sigma_{\text{initial}, r}^t
	\label{eq_sigma}
\end{equation}
where $r \in \{long, lat\}$. $\textit{v}_{max}$ describes the highest velocity in the dataset and $\sigma_{\text{max}, r}^t$ sets a maximum value for the standard deviation in $r$ at time $t$. We rotate the Gaussian kernel by $\Theta$ to align it with the heading direction of a road user at time $t$ in the grid. In practice, the standard deviation in the driving direction $\sigma_{long}$ is set to be higher than in the lateral direction $\sigma_{lat}$. The purpose of the Gaussian kernel in the ground truth images is explained in more detail in the loss function subsection.

\subsection{Network Architecture}
An overview of the network architecture is illustrated in Fig. \ref{fig_caspnet}. It consists of four key components: A \textit{trajectory encoder}, \textit{map encoder}, \textit{skip connections} and a \textit{trajectory decoder}. When designing this architecture, we considered the problem of predicting multiple different road users in a scene at the same time. It is a challenging task due to highly complex scenarios with various numbers and different types of road users who have different maneuverabilities and can impact each other's motion. Moreover, the static context needs to be considered simultaneously.

\begin{figure}[t]
	\centering
	{\includegraphics[width=0.45\textwidth]{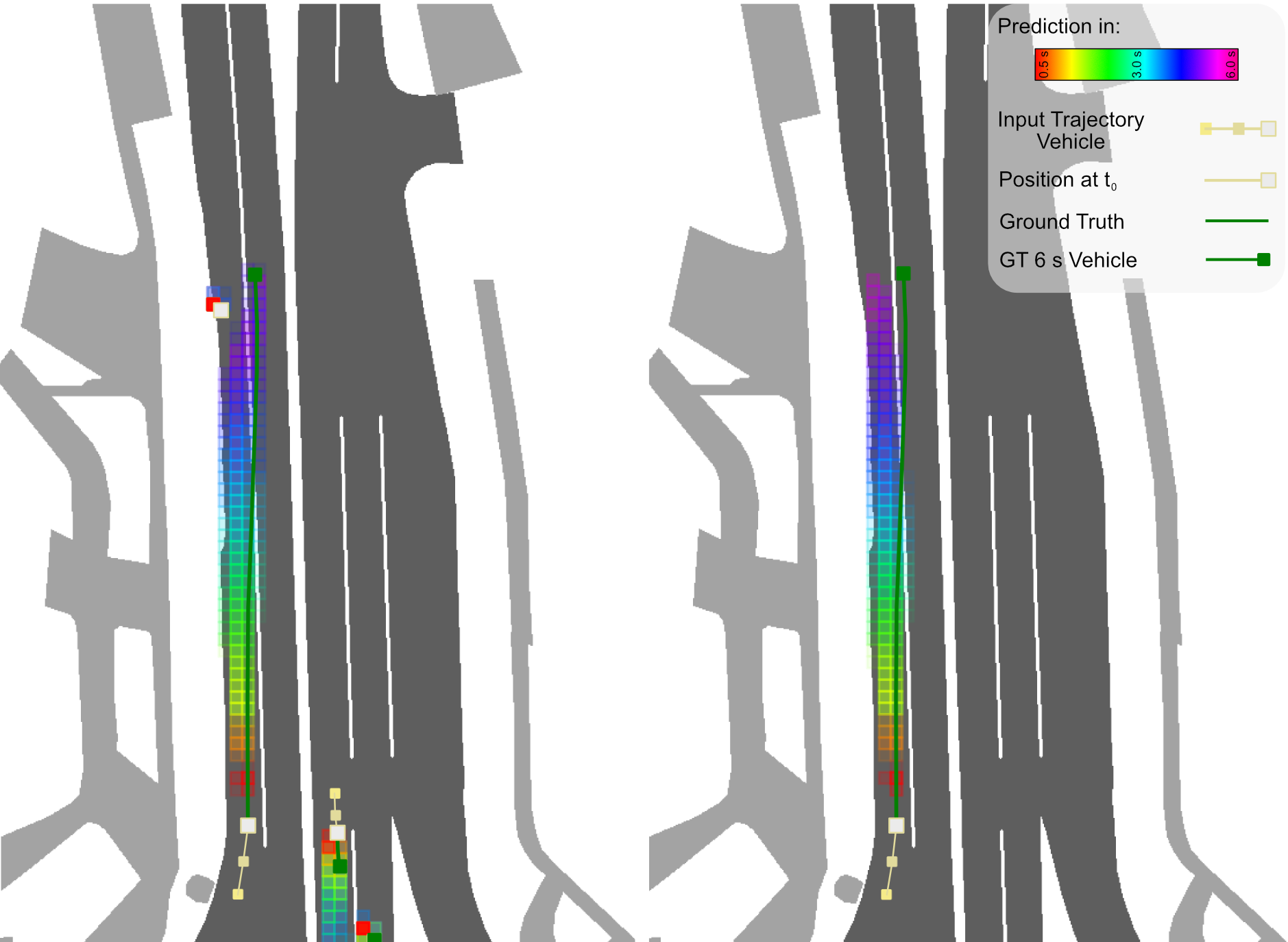}}
	\caption{Prediction of the same scene using different inputs for the same network. On the left, the predicted future positions for the vehicle in the left lane show a possible lane change at a higher time horizon to avoid a collision with the stationary vehicle. On the right, by artificially removing other vehicles, one may see that the predicted future positions of the same vehicle are within its own lane. There is no reason for it to change the lane.}
	\label{fig_interaction_example}
\end{figure}
Fig. \ref{fig_interaction_example} demonstrates a simple interaction between a moving vehicle and a stationary one in the same lane. This interaction happens spatially at a far range, which also corresponds to a higher future time horizon. There are various interactions among the road users, not only at far range and not only among vehicles. This leads to the idea of applying feature pyramid networks (FPN) throughout the CASPNet architecture, allowing features covering different-sized receptive fields to flow through the network. Not only are these interactions captured spatially, but by temporal filtering inside the skip connections, CASPNet can capture the temporal correlations among the road users' motions. In the end, CASPNet is able to learn complex interactions from real-world traffic scenes. This capability will not only be statistically demonstrated in the evaluation section but also visually. Thus, the viewer can intuitively interpret the prediction results.

\subsubsection{Trajectory Encoder}
The trajectory encoder consists of multiple CNN blocks, and each block consists of a 2D convolution layer followed by batch normalization and a max pooling layer (except for the first block). The series of trajectory images are used as input for the trajectory encoder. Each image, containing the features for one time step, is encoded through multiple CNN Blocks. The trajectory encoder's weights are shared for the encoding of the input grids at different time steps $T_i$.

\subsubsection{Map Encoder}
When driving fast, the driver needs to observe the road far ahead, whereas a slow walking pedestrian may pay more attention to his close by surroundings. Due to this reason, the map is also encoded through a FPN, and features of all pyramid levels are passed through the skip connections to the decoder. On the rasterized HD map, we sequentially apply two Gabor Convolution Networks (GCN) \cite{gabor} and CNN Blocks. We use GCN to improve the resistance to changes in orientation and scale of the features \cite{gabor}. At the end of the second GCN, we use a max pooling layer. The CNN Blocks have the same architecture as the ones from the trajectory encoder.

\subsubsection{Skip Connections}
The trajectory input images at different time steps are encoded independently, to learn the temporal correlation among them, we use ConvLSTM layer inside the skip connections. Since the rasterized HD map is fixed to the time step $t_0$, the ConvLSTM layer is applied before the map encoder. If each input image has its own map, the ConvLSTM layer can be easily positioned behind the map encoder, thus the dynamic of the map can also be filtered. After the ConvLSTM, the trajectory feature maps are concatenated with the ones from the map encoder at the respective pyramid level.

\begin{figure}[t]
	\centering
	{\includegraphics[width=0.45\textwidth]{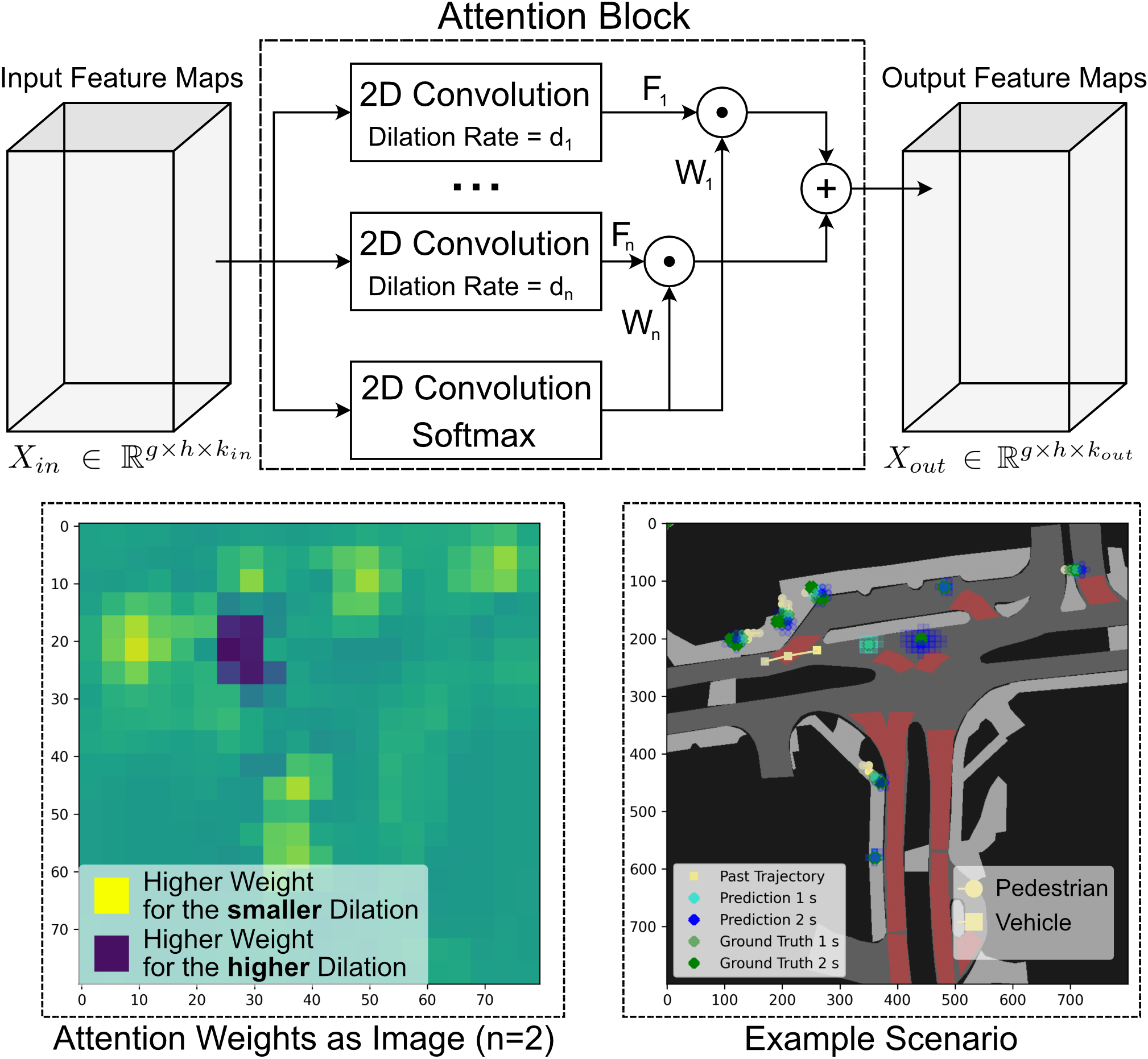}}
	\caption{On the top: Architecture of the Attention Block for learning the correlation among road users' motion and their dynamic and static context at a variable range at individual pixels in the same feature map. On the bottom left: Predicted attention weights for the scenario shown on the bottom right.}
	\label{fig_dilated_conv}
\end{figure}

The FPN in the trajectory and map encoder helps to capture features with different receptive fields spatially. Each level of pyramid features passing through the skip connection still captures the movements of different types of road users, which can have different dynamics and maneuverabilities. The attention block is introduced to provide variable receptive fields at pixel-level for the feature map passing through a skip connection. The Attention Block architecture is illustrated in Fig. \ref{fig_dilated_conv}.

We propose an approach to approximate features that cover individual-sized receptive fields in the same output feature map by applying an attention mechanism. The input feature map is defines as $X_{in} \in \mathbb{R}^{g \times h \times k_{in}}$ and the output feature map as $X_{out}\in \mathbb{R}^{g \times h \times k_{out}}$, where $g \text{ and } h$ define the spatial size and $k_{in} \text{ and } k_{out}$ the number of input and output channels. Multiple 2D convolutions with different dilation rates $d \in D$ on the input feature map are used, where $D$ is a set of $n$ different dilation rates. With an additional 2D convolution layer, the network is able to learn attention weights $W \in \mathbb{R}^{g \times h \times n}$ from the same input which are then normalized by applying a softmax function over the channel dimension:
\begin{equation}
	W(X_{in}) = \text{Softmax}(\text{Conv}(X_{in})).
	\label{eq_attention}
\end{equation}
It is important to note that the kernel size of the convolution in Eq. \ref{eq_attention} matches the receptive field of the convolution with the highest dilation rate. The resulting feature maps from the dilated convolutions are multiplied element-wise with the corresponding attention weight:
\begin{equation}
	\resizebox{0.91\hsize}{!}{%
		$	{X_{out}(X_{in}) = \text{Conv}_{d_1}(X_{in})\odot W_1 + ... +  \text{Conv}_{d_n}(X_{in}) \odot W_n}$%
	}
	\label{eq_individual_receptive_field}
\end{equation}
where $Conv_{d_i}$ is a 2D convolution with the $i$-th value in $D$ as  dilation rate. As a result, each pixel in a feature map can be processed with different weighted receptive fields. 

\begin{figure*}[t]
	\centering
	{\includegraphics[width=0.95\textwidth]{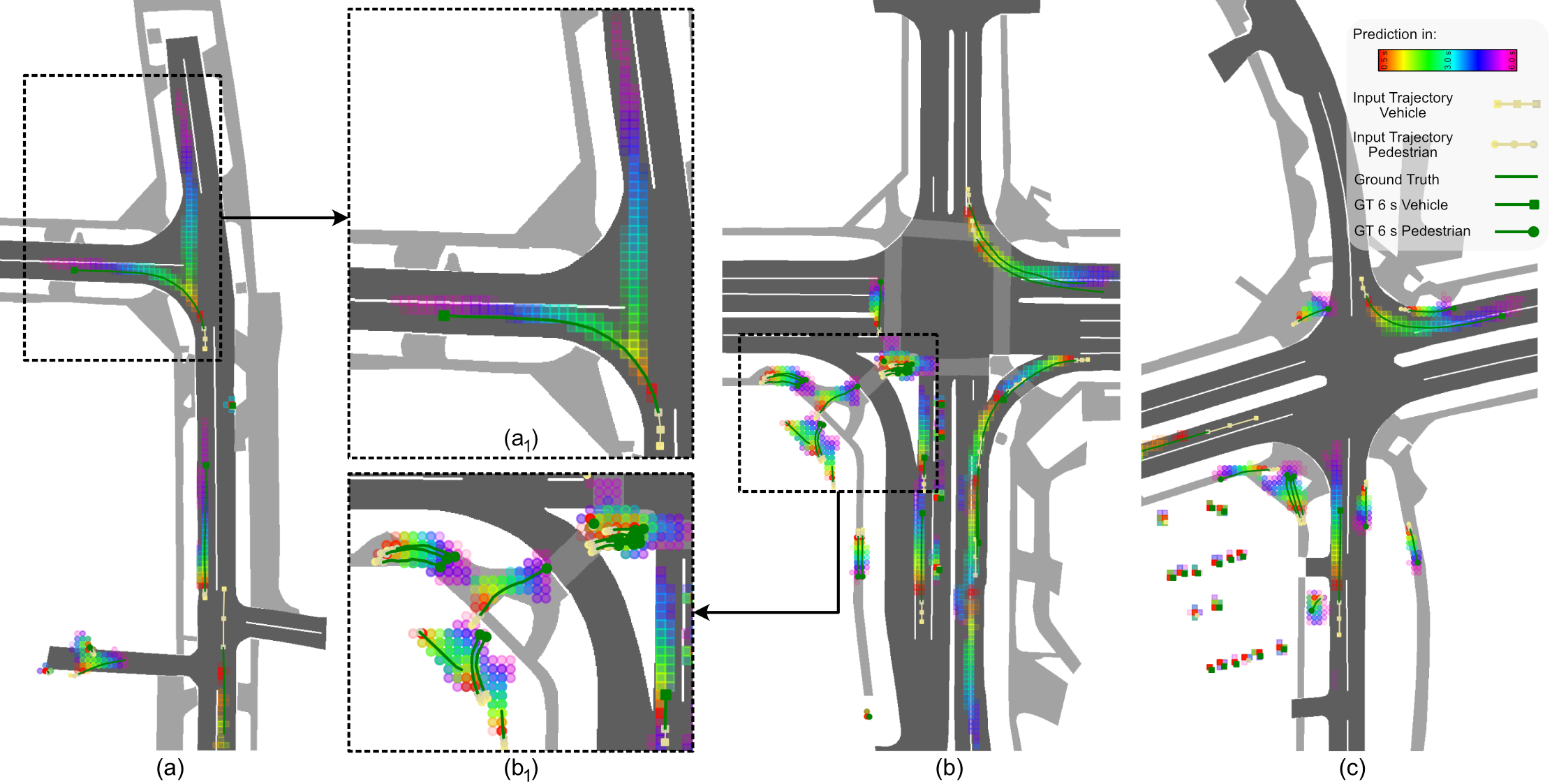}}
	\caption{Qualitative examples of CASPNet scene prediction on the nuScenes dataset. The series of predicted grids from different future time steps are overlapped and visualized in distinct colors from red (prediction in 0.5 s) to magenta (6 s). The Ground Truth is visualized in green. We show the prediction for three challenging scenes (a)-(c): ($a$) Multi-road scene, ($a_1$) multi-modal prediction at a T-intersection, ($b$) intersection with complex interaction scenario, ($b_1$) crowded pedestrian scenario and ($c$) intersection scene.}
	\label{fig_scene_quality}
\end{figure*}

\begin{figure}[!b]
	\centering
	{\includegraphics[width=0.4\textwidth]{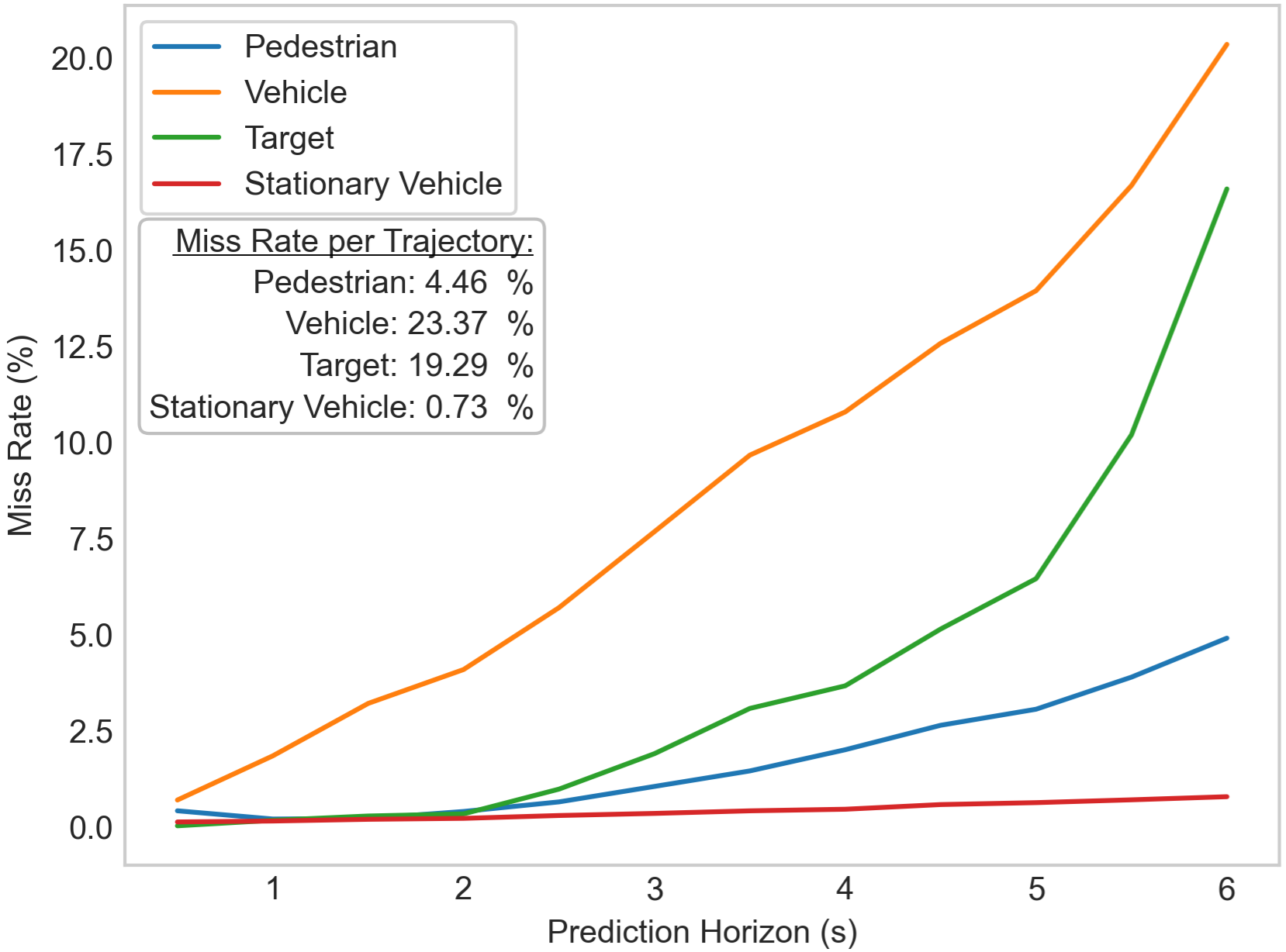}}
	\caption{Quantitative results of CASPNet scene prediction on the nuScenes dataset. We report the miss rate per time step for the classes pedestrian, vehicle, target and stationary vehicle. For said classes, we also calculate the miss rate per trajectory.}
	\label{fig_scene_quantitativ}
\end{figure}

\subsubsection{Trajectory Decoder}
In the trajectory decoder, we use residual upsampling introduced in \cite{wojna2019devil} to sample the feature maps back up to the defined output resolution. The residual upsampling block consists of parallel bilinear upsampling and transpose convolution layer. The resulting feature maps are added and then concatenated with the feature map from the next higher pyramid level. A residual layer similiar to Inception-ResNet-B in \cite{Szegedy2017} is applied afterwards using multiple parallel convolution with different kernel sizes ($(1, 1)$, $(7, 1)$, $(1, 7)$, $(3, 3)$). 

The final output layer is a ConvLSTM which takes the output feature map from the residual upsampling block and then iteratively propagates the hidden state. At each iteration $o \in \{1,..,N\}$, it outputs a prediction at $t_o$ future time step. This allows us to predict at flexible time horizons. In addition, the decoder path only has to be executed once for each sample and there is no need for multiple output heads.

\subsection{Loss Function}
We use a distance aware focal loss for the classification task inspired by CornerNet \cite{Law_2018_ECCV}:
\begin{equation}
	\resizebox{0.91\hsize}{!}{%
		$	{L_{tc} = {-1 \over {E_{tc}+1}} \sum_{u=0}^{U} \sum_{v=0}^{V}
			\begin{cases}
				\alpha (1-p)^\gamma log(p) & \text{if $Y_{tuvc}$ = 1}\\
				(1-\alpha) (1-Y_{tuvc})^\beta p^\gamma log(1-p) & \text{otherwise}
		\end{cases}}$%
	}
	\label{eq_class_loss}
\end{equation}
where p stands for $O_{tuvc}$, the predicted pixel of class $c$ and time step $t$. $E_{tc}$ describes the total number of objects from class $c$ at time step $t$. The Gaussian kernel around positive labels in the GT images reduces the loss nearby the GT position of an object. Thus, false positive predictions that are close to a positive label, which are still plausible predictions, are not as strongly penalized as predictions that are further away. Unlike \cite{Law_2018_ECCV}, we use $\alpha$ proposed by \cite{focalLoss} to cope with the high imbalance between positive and negative labels. We set $\alpha=0.25$, $\beta=4$ and $\gamma=2$ based on hyperparameter tuning.
The class loss for all time steps is summed up:
\begin{equation}
	\resizebox{0.41\hsize}{!}{%
		$	{ L_{\text{class}} = \sum_{c=1}^{|C|} \sum_{t=1}^{N} w_{c} L_{tc}
		}$%
	}
	\label{eq_combined_class_loss}
\end{equation}
where $C = \{\text{target, vehicle, pedestrian}\}$ and $w_c$ is a predefined weight for class $c \in C$. The offset prediction is shared for all classes. Its loss is defined as the sum of squared errors between prediction and GT over all pixels from all grids:
\begin{equation}
	\resizebox{0.91\hsize}{!}{%
		$	{
			L_{\text{offset}} = \sum_{t=1}^{N} \sum_{u=0}^{U} \sum_{v=0}^{V}
			\begin{cases}
				(O_{tuv\delta_{u}} - Y_{tuv\delta_{u}})^2 + \\ (O_{tuv\delta_{v}} - Y_{tuv\delta_{v}})^2 & \text{if $ \sum_{c=1}^{]C]} Y_{tuvc} = 1 $}\\
				0 & \text{otherwise}
		\end{cases}}$.%
	}
	\label{eq_offset_loss}
\end{equation}
Finally, our loss function is given as the sum of the classification and offset loss divided through the number of predicted time steps:
\begin{equation}
	Loss =  {1 \over N} (L_{\text{class}} + L_{\text{offset}}).
	\label{eq_joint_loss}
\end{equation}

\subsection{Implementation Details}
The models were trained on an Nvidia GeForce RTX 2080 Ti GPU. The batch size is 16. The Adam optimizer with an initial learning rate of 0.0001 is used. Due to the rather small dataset size, we made use of data augmentation methods to cope with overfitting. During training, we randomly rotate the samples between 0 and $2\pi$ and add a random global translation [-3, 3] so that the current position of the target or ego object is not always at the same location on the grid. Inside the Attention Block, we use a the dilation rates $D=\{1, 2\}$.

\section{EVALUATION SCENE PREDICTION}
\subsection{Dataset}
For testing, the publicly available dataset nuScenes \cite{Caesar2020} is used which contains a high amount of labeled vehicle and pedestrian trajectories in urban scenarios. In total, there are 1000 scenes recorded in Boston and Singapore each 20 seconds long. In addition, HD maps are provided for the regions containing rich information about the static environment such as driveable areas, lane boundaries and crossing areas. The task of the nuScenes prediction benchmark is to predict the future trajectory of a target vehicle for the next 6 seconds at 2 Hz, given a maximum of 2 seconds of the past trajectory as well as static and dynamic context information. In our tests, we only utilize one second of the past trajectory as input.

\begin{figure*}[t]
	\centering
	{\includegraphics[width=0.95\textwidth]{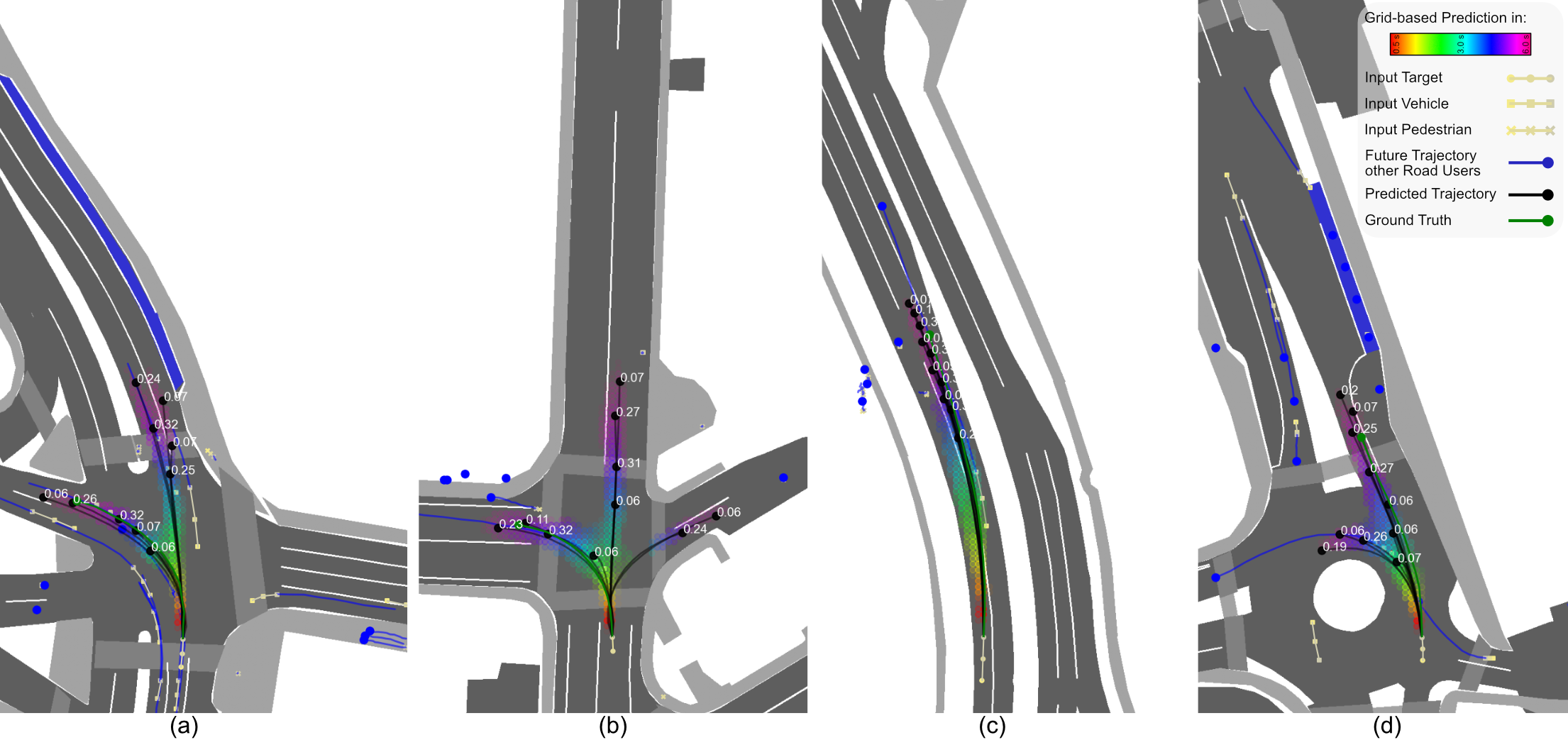}}
	\caption{To facilitate comparisons with other state-of-the-art approaches, we tested for single target prediction and show qualitative examples on the nuScenes dataset. The predicted grids for each time step are overlapped and visualized in different colors (from red prediction in 0.5 s to magenta in 6.0 s). The $K$ extracted trajectories from the grid-based prediction are visualized in black. The corresponding probability of each extracted trajectory is illustrated at the last position. Four different scenarios are purposely picked to demonstrate the performance of CASPNet. (a) Prediction and extracted trajectories for a vehicle in front of a irregular intersection: The extracted trajectories based on CASPNet output overlap the left turn lane and two straight forward lanes. Given the input trajectory and its location in front of the intersection, these trajectories are feasible and also possible futures. (b) Prediction and extracted trajectories for a vehicle in front of a four-way intersection with no designated turn lanes: Three directions are predicted, with the left-turn being more likely than the other options. (c) Prediction and extracted trajectory on a two lanes curved road: A lane change is correctly predicted since the target's lane is occupied by a pedestrian and a stationary vehicle. (d) Prediction and extracted trajectory of a vehicle just entering a round-about: Following and exiting the roundabout are both predicted.}
	\label{fig_single_quality}
\end{figure*}

\subsection{Metrics}
We evaluate the performance of CASPNet with the \textbf{miss rate} (MR) metric. A future GT position is missed if the maximum L2 distance to any predicted occupied point is larger than 2 meters. The MR per time step is the fraction of misses to the number of GT trajectory points in the ROI. A trajectory is classified as a miss if any of its time steps is missed. 

The track-based miss rate is defined as the fraction of missed trajectories to the number of GT trajectories. CASPNet predicts future positions of all road users using grids and does not have a fixed number of modalities. For this reason, we evaluate the scene prediction on the most likely points extracted from the grid over a threshold and not over a predefined number of modes.

\subsection{Qualitative results}
Qualitative results are shown for the scene prediction in Fig. \ref{fig_scene_quality}. We focus on examples of complex scenes which include vehicles turning, driving straight, or breaking in front of standing vehicles, and pedestrians crossing the road, following the sidewalk, moving outside of the sidewalk, or avoiding collision with vehicles. The predictions show a distribution-like structure, and the GT trajectory overlaps very well with the predicted points. This does not only visually demonstrate the performance of the CASPNet in an easily interpretable manner but also visually shows the possible variations of the future trajectory.

\subsection{Quantitative Results}
Quantitative results of CASPNet scene prediction are shown in Fig. \ref{fig_scene_quantitativ}. To our best knowledge, there are no other scene prediction approaches on nuScenes which have a grid-based output representation, thus we cannot compare with other approaches. For this reason, we also evaluated CASPNet for single target prediction. By post-processing of the predicted grid-based output, track-based trajectories are extracted so that a comparison with state-of-the art approaches is possible. More details are provided in the next section.

\subsection{Attention Block}
To help understand the effectiveness of our Attention Blocks, we visualized the attention weights and compared them to a given scenario. A vehicle, which moves at a faster velocity, should have a greater receptive field. This corresponds to the CNN layer with a higher dilation rate in the Attention Block. On the bottom of Fig. \ref{fig_dilated_conv}, the left figure visualizes the pixel level weights of the CNN layers with different dilation rates for the scene on the right. The figure on the right shows the current positions of vehicles and pedestrians as well as the final prediction and GT. Comparing these two plots, it is clear to see that pixels close to the vehicle position generally prefer the CNN layer with a greater dilation rate. At the pedestrians' positions, it is the other way around, confirming our initial expectations. It is worth noting, that although there are no pedestrians on the right side of the sidewalk, there is still a higher weight on the smaller dilation, indicating a good understanding of the static context as a vehicle driving there at high speed is rather unlikely.

\section{EVALUATION SINGLE TARGET PREDICTION}
For a fair comparison with state-of-the-art trajectory prediction systems, which focus on the prediction of single road user's trajectory, we also evaluate CASPNet as a single target predictor.

\subsection{Single Target Prediction}
Instead of predicting all road users in the scene, one can also only pick one specific vehicle as target and make a target specific prediction. This is done by assign the chosen target vehicle the type $c_{target}$ in Eq.  \ref{eq_input_features}, and keep the other road users' types unchanged. In the output in Eq. \ref{eq_output_features} and GT only $c_{target}$ and the in-pixel offset are used. Eq. \ref{eq_combined_class_loss} simplifies to only the sum over the time steps for the class target. With these minor changes and the same network architecture, CASPNet can predict the motion of a single target. Fig. \ref{fig_single_quality} demonstrates some single target prediction examples.

\begin{table*}[t]
	\caption{Results on the nuScenes prediction challenge test split}
	\centering
	\resizebox{0.6\hsize}{!}{%
		\begin{tabular}{ l | r r | r r | r r  | r}
			
			\textit{} & \multicolumn{2}{c} {K=10} & \multicolumn{2}{c} {K=5}  & \multicolumn{2}{c} {K=1} & \textit{} \\
			
			Method & \textit{MR} & \scriptsize \textit{min}ADE & \textit{MR} & \scriptsize \textit{min}ADE & \scriptsize \textit{min}FDE & \scriptsize \textit{min}ADE & \textit{OR} 
			\\  
			\hline
			Constant Velocity \cite{Phan-Minh2020} & 0.91 & 4.61 & 0.91 & 4.61 & 11.21 & 4.61 & 0.14 \\
			Physics Oracle \cite{Phan-Minh2020} & 0.88 & 3.70 & 0.88 & 3.70 & 9.09 & 3.70 & 0.12 \\
			CoverNet \cite{Phan-Minh2020} & 0.64 & 1.92 & 0.76 & 2.62 & 11.36 & - & 0.13 \\
			\hline
			WIMP \cite{Khandelwal2020} & \textbf{0.43} & 1.11 & \textbf{0.55} & 1.84 & 8.49 & - & 0.04 \\
			MHA-JAM \cite{Messaoud2020} & 0.45 & 1.24 & 0.59 & 1.81 & 8.57 & 3.69 & 0.07 \\
			cxx \cite{Luo2020} & 0.60 & 1.29 & 0.69 & 1.63 & 8.86 & - & 0.08 \\
			Noah & 0.62 & 1.37 & 0.69 & 1.59 & 9.23 & - & 0.08 \\
			P2T \cite{deo2021trajectory} & 0.46 & 1.16 & 0.64 & 1.45 & 10.50 & - & 0.03 \\
			GOHOME \cite{gilles2021gohome} & 0.47 & 1.15 & 0.57 & 1.42 & \textbf{6.99} & - & 0.04 \\
			Autobot & 0.44 & \textbf{1.03} & 0.62 & \textbf{1.37} & 8.19 & - & \textbf{0.02} \\
			\hline
			CASPNet (Ours) & \textbf{0.43 }& 1.19 & 0.60 & 1.41 & 7.27 & \textbf{3.16} & \textbf{0.02} \\
	\end{tabular}}
	\label{tab_nuscenes_result}
\end{table*}

\subsection{Trajectory Extraction from Grid-based Output}
\begin{figure}[h]
	\centering
	{\includegraphics[width=0.45\textwidth]{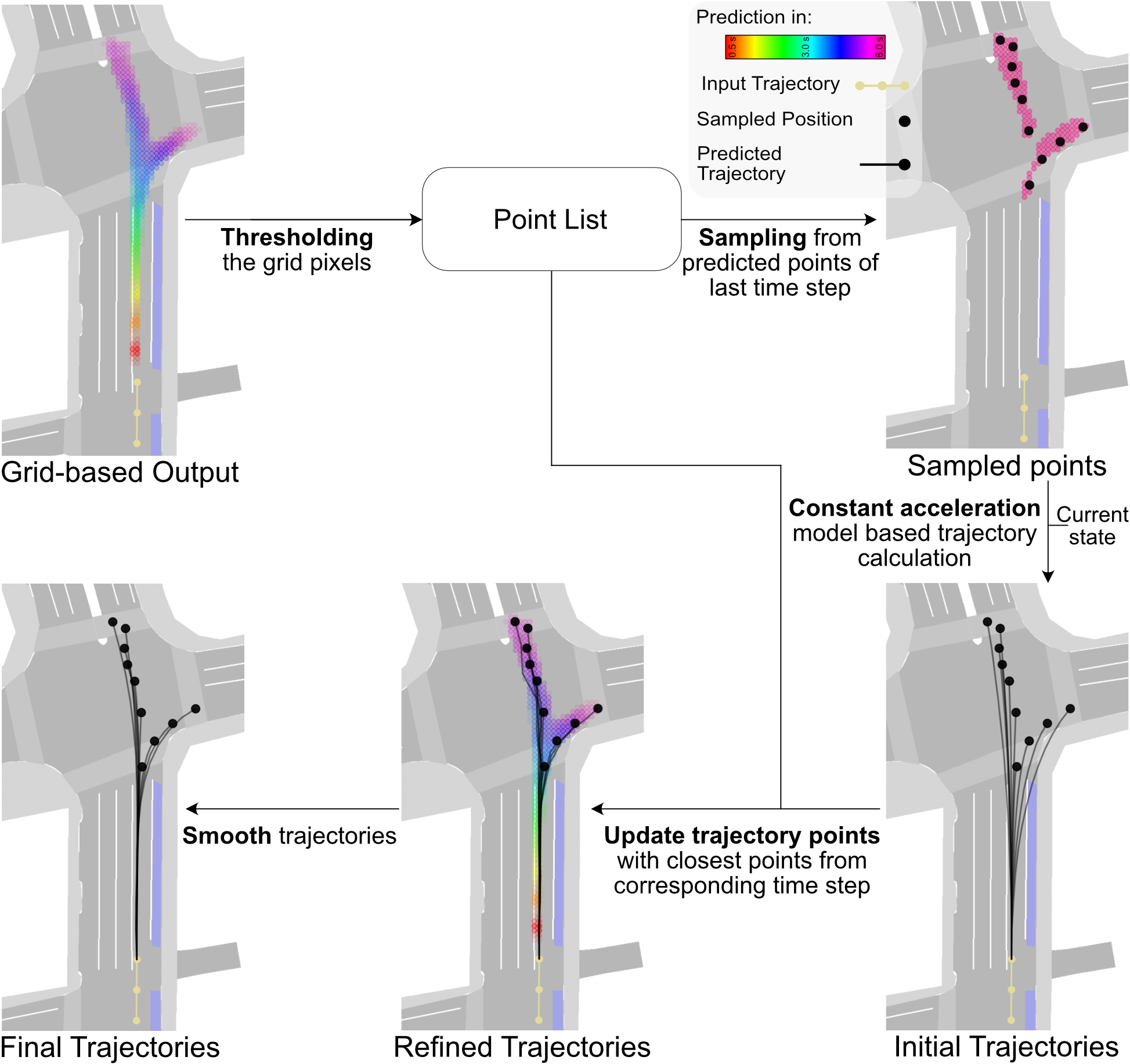}}
	\caption{Process of extracting trajectories from the grid-based prediction of CASPNet. For visualization purpose, the predictions are overlapped with the HD map.}
	\label{fig_trajectory_reconstruction}
\end{figure}

To be able to compare with other trajectory prediction methods, a track-based output structure is necessary. For trajectory extraction, we use the output of the single target prediction CASPNet. The extraction process is visualized in Fig. \ref{fig_trajectory_reconstruction}.The grid-based output for each time step is transformed into a point-wise structure by adding the predicted offset to each pixel position $(u, v)$ over a threshold and writing them to a vector of points. By applying a non-maximum-suppression (NMS) on the points from the last prediction horizon, using a dynamic thresholding distance between sampled points, we extract the top five most probable modalities. With farthest point sampling, we add another five diverse points. Using a constant acceleration model, we calculate initial future trajectories from the current road user state to the sampled last positions. The initial trajectories are updated with the closest extracted points from the corresponding time steps. For improvement of the temporal consistency, we smooth the resulting trajectories. The probability of a trajectory is defined through the probability of the sampled last position from the predicted grid.

\subsection{Metrics}
To evaluate and compare our single target prediction with state-of-the-art methods the most commonly applied track-based metrics in motion prediction are used. We use \textbf{minADE} and \textbf{minFDE} and report these metrics for different numbers of modes (K). In addition, the track-based MR is used. The \textbf{off-road rate} metric (OR) describes the ratio of predicted trajectories laying not entirely in the driveable area of the map to the total number of predicted trajectories. It is an indicator of the adequacy of the predicted trajectories to the static context in a scene.

\subsection{Qualitative Results}
In Fig. \ref{fig_single_quality}, single target prediction and extracted trajectories are visualized. Four challenging scenarios are demonstrated: Irregular intersection, 4-way intersection, multi-lane curved road with a parked vehicle, and a round-about.

\subsection{Quantitative Results} We compare the result from our method on the nuScenes prediction challenge to the baseline methods Constant Velocity, Physics Oracle and CoverNet \cite{Phan-Minh2020} and to state-of-the-art methods in Tab. \ref{tab_nuscenes_result}. The results in the benchmark are ranked by ${minADE(K=5)}$ where we are able to rank second overall. In the \textit{OR} metric we tie with the best approach indicating a good understanding of the static context from our model.

\section{SUMMARY AND FUTURE WORKS}
In this work, a framework for jointly predicting the motion of all road users in a scene was presented. The proposed approach solves the challenging task of predicting complex scenes with a variable number of road users that may have different maneuverabilities. Using grid-based input and output data structures, the network is able to predict multi-modal future motion of all the road users in the scene. A bonus is the fixed computational cost, which alone could be an attractive feature. The proposed Attention Block learns interactions between road users as well as their static contexts at diverse ranges.
Quantitative and qualitative evaluations on the nuScenes dataset show that our approach reaches state-of-the-art results. 

Depending on concrete applications, some may require the predicted trajectories to be extracted from the grid map output, e.g. for simulation where only a set of trajectories may occur, which need to be jointly plausible and scene consistent. Thus, we plan to extend the trajectory extraction for all the agents. At the same time, we argue that our grid map output could directly serve as the input for collision avoidance modules, in which the trajectories given in separate parametric representation may not be necessary. Also, depending on the available data, we would like to extend our static context to cover traffic signs, traffic lights, etc. Another exciting direction is to extend CASPNet for a joint prediction and path planing task.

\addtolength{\textheight}{-12cm}   


\bibliographystyle{IEEEtran}
\bibliography{./IEEEtranBST/IEEEexample}

\end{document}